\documentclass[letterpaper]{article} 
\usepackage{aaai24}  
\usepackage{times}  
\usepackage{helvet}  
\usepackage{courier}  
\usepackage[hyphens]{url}  
\usepackage{graphicx} 
\usepackage{natbib}  
\usepackage{caption} 
\usepackage{algorithm}
\usepackage{algorithmic}

\usepackage{mathrsfs}

\usepackage{newfloat}
\usepackage{listings}
\DeclareCaptionStyle{ruled}{labelfont=normalfont,labelsep=colon,strut=off} 
\floatstyle{ruled}
\newfloat{listing}{tb}{lst}{}
\floatname{listing}{Listing}
\title{Uncertainty Quantification in Detecting Choroidal Metastases on MRI via Evolutionary Strategies}

\author{ Bala McRae-Posani\textsuperscript{\rm 1}, Andrei Holodny\textsuperscript{\rm 1}, Hrithwik Shalu\textsuperscript{\rm 2}\equalcontrib, Joseph N Stember\textsuperscript{\rm 1}\equalcontrib }
\affiliations{

    \textsuperscript{\rm 1}Memorial Sloan Kettering Cancer Center, Department of Radiology\\

    1275 York Avenue, New York, NY 10065\\
    mskcc.org\\

    \textsuperscript{\rm 2}Indian Insitute of Technology Madras, Department of Aerospace Engineering\\
    IIT P.O, Chennai, Tamil Nadu, India, 600036\\
    iitm.ac.in

}

\usepackage{bibentry}

\begin{document}

\maketitle

\begin{abstract}
Uncertainty quantification plays a vital role in facilitating the practical implementation of AI in radiology by addressing growing concerns around trustworthiness. Given the challenges associated with acquiring large, annotated datasets in this field, there is a need for methods that enable uncertainty quantification in small data AI approaches tailored to radiology images. In this study, we focused on uncertainty quantification within the context of the small data evolutionary strategies-based technique of deep neuroevolution (DNE). Specifically, we employed DNE to train a simple Convolutional Neural Network (CNN) with MRI images of the eyes for binary classification. The goal was to distinguish between normal eyes and those with metastatic tumors called choroidal metastases. The training set comprised 18 images with choroidal metastases and 18 without tumors, while the testing set contained a tumor-to-normal ratio of 15:15.\\

We trained CNN model weights via DNE for approximately 40,000 episodes, ultimately reaching a convergence of 100\% accuracy on the training set. We saved all models that achieved maximal training set accuracy. Then, by applying these models to the testing set, we established an ensemble method for uncertainty quantification.The saved set of models produced distributions for each testing set image between the two classes of normal and tumor-containing. The relative frequencies permitted uncertainty quantification of model predictions. Intriguingly, we found that subjective features appreciated by human radiologists explained images for which uncertainty was high, highlighting the significance of uncertainty quantification in AI-driven radiological analyses.
\end{abstract}

\section{Introduction}

\subsection{Radiology artificial intelligence (AI) requires Uncertainty quantification (UQ)}

It is widely appreciated that for artificial intelligence (AI) to gain traction in real-life clinical deployment, single-point predictions must be contextualized by uncertainty quantification (UQ). UQ can help ensure that AI predictions are trustworthy and reliable, helping to overcome clinician hesitation about using them in practice \cite{faghani2023quantifying,faghani2022mitigating,mccrindle2021radiology}. By incorporating UQ, AI researchers can produce responsble AI better positioned to benefit patients directly. Although not phrased in those terms, the importance of UQ for patient management is reflected in the fact that several radiology report lexicons have been designed expressly to convey the certainty or uncertainty of diagnoses \cite{panicek2016sure,kendall2017uncertainties}. 

\subsection{UQ must work in settings of sparse data}

Despite UQ’s key role in responsible AI, there has hitherto been no practical way to evaluate uncertainty for small data scenarios. Practical radiology AI requires being able to learn in the small data regime since obtaining and labeling images is time-consuming and tedious and excludes rare conditions or presentations. 

Hence, the radiology AI research community needs small data UQ. In this work, we will demonstrate a way to quantify uncertainty for classifying small sets of radiological images. 

\subsection{We quantified uncertainty using a small data training approach based on evolutionary strategies}

In particular, we will show that UQ with small training sets is feasible when we use a small data training approach called Deep Neuroevolution (DNE) \cite{stember2022deep,stember2023direct,purkayastha2022deep}. We have chosen to use DNE because it has been shown to achieve high training and testing set accuracies for small numbers of brain MRIs, and can generalize to heterogeneous data. 

We studied eye MRI exams to demonstrate how researchers can embed UQ into DNE's small-data AI approach. In particular, we performed binary classification between normal eyes and those with metastatic lesions, called choroidal metastases. This provides a nice illustration of computing and studying UQ when only limited data are present.

\section{Methodology}

From recent literature, it has been established that Evolutionary Strategies (ES)-based parametric optimization of Neural Networks are scalable alternatives in the regime of Data-Efficient Deep Learning\cite{stember2023direct, such2017deep}. In this work we employ an ES based optimization strategy called Deep Neuroevolution (DNE) to form Deep ensembles. DNE has the valuable property of permitting the semiautomated acquisition of an ensemble of different, but all valid, models. We can quantify epistemic uncertainty in unseen testing set data using the model ensembles. A straightforward advantage for gradient-free optimization methods compared to autograd\cite{ketkar2021automatic}-/gradient descent-based methods is the flexibility granted on the constraint/objective function used. For autograde-based deep learning, the objective function is the loss of the output with respect to a known human-annotated label or target value.  For our purposes, we form a scalar reward (fitness) based on the sum of True Positives and True Negatives that acts as our evaluation metric.

\subsection{A brief discussion of Evolutionary Strategies (ES)}

The form of ES used throughout this study can be simplified into three major components. To perform parametric optimization of a deep neural network, firstly, a population of models with size 2p is formed from a reference model by asynchronous perturbations of the model weights. Suppose the weights of the reference model at some iteration standpoint $t$ is $w_t$, then $p$ pairs of perturbed models are formed as the positive/negative perturbations of the model weights. More formally, the weights of a pair of perturbed models ($\pm p_i$) are obtained by perturbing the reference model weights (at iteration $t$) with a random vector ($\epsilon_i$) as shown in Equation~\ref{weight_perturb_eq}, where $\mu$ is the magnitude factor of the perturbations:

\begin{equation}
    w_{p_i}^{\pm} = w_t \pm \mu \epsilon_i.
    \label{weight_perturb_eq}
\end{equation}

\noindent
Note that the values of perturbation vector $\epsilon_i$ for all our purposes are randomly sampled from a standard normal distribution. Secondly, we evaluate the perturbed models with a given training set ($\mathcal{D}_T$), and the metric formed ($R_i^\pm$) is normalized as per the scheme shown in Equation \ref{Eq_1_3}. 

\begin{equation}
    ^N R_i^\pm = \frac{\max \left(0,\; \log_2(p + 1) - \log_2( R_i ^\pm + 1)\right)}{\sum_{i=0}^{2p-1} \max\left(0,\; \log_2(p + 1) - \log_2( R_i^\pm + 1)\right)} + \frac{1}{2p}
    \label{Eq_1_3}
\end{equation}

\noindent
\textit{Finally}, we update the reference model with gradients computed per the SPSA~\cite{spall1992multivariate} scheme. More formally the gradient estimates formed by evaluating on $p$~pairs of perturbed models with their respective performance metrics~($^N R_i^\pm$) is as shown in Equation~\ref{spsa}.
\begin{equation}
 \nabla_{w_t} \mathcal{R}  = \frac{1}{p} \sum_{i=1}^{p} \frac{^N\mathcal{R}_{i}^{+} (w_t + \mu \epsilon_i) - ^N\mathcal{R}_{i}^- (w_t - \mu \epsilon_i)}{2 \mu}
\label{spsa}
\end{equation}

\noindent
A noteworthy advantage of the optimization scheme is the asynchronous nature of the gradient updates and their scalable nature in terms of the population size $p$. That is, relative exploration of the parameter space could be improved with a large enough population size \cite{assunccao2019fast}, enabling greater diversity regarding neuron pathways in the formed models.

\subsection{Estimating prediction uncertainty using deep ensembles}

In this section, we outline the method used for quantifying the epistemic uncertainty of models trained using the ES-based optimization scheme. We base our method on the fact that there is diversity in the population of models formed at each iteration, including after training has reached convergence. DNE's convergence properties enable a semi-automated and reliable ensemble-based UQ scheme. DNE has been shown to converge monotonically to high training and testing set accuracies \cite{stember2022deep,stember2023direct,stember2023deep,purkayastha2022deep} DNE converges monotonically toward high training and testing set accuracies. Hence, after convergence, all models both:
\begin{itemize}
    \item Valid because they are highly accurate. and
    \item Unique, because of the random perturbations applied to network weights at each step of DNE.
\end{itemize}
This permits a straightforward, semiautomated, and reliable method for UQ:
\begin{enumerate}
    \item Train the CNN with DNE until convergence has been reached.
    \item Once some minimum number $N_{conv}$ of training steps/generations have transpired since convergence, training can be terminated.
    \item Simply collect the last $N_{conv}$ model weights. $N_{conv}$ represents the post-convergence training iterations, and represents the desired number of ensemble members for UQ. In general, higher $N_{conv}$ is preferred due to better statistical sampling. An advantage of DNE-based UQ is that we can readily collect as many models as desired to obtain sufficient statistical sampling for UQ, since all we need to do is wait for convergence and then collect/save models.
    \item Once we have our ensemble of models $\Bigl\{ \mathscr{F}_i  \Bigr\}_{i=1}^{N_{conv}}$, we can compute a distribution of class predictions $C$ for each testing set image $I$: $\Bigl\{ C_i  \Bigr\}_{i=1}^{N_{conv}}$ = $\Bigl\{ \mathscr{F}_i (I) \Bigr\}_{i=1}^{N_{conv}}$. Of note, here we have a binary classifier, so $C$ can be either metastatic tumor ($\mathscr{T}$) or normal/tumor-free ($\mathscr{N}$).
    \item Quantify the uncertainty by first computing its inverse, the Shannon entropy, $S$. For the two classes, tumor ($\mathscr{T}$) and normal ($\mathscr{N}$), this becomes: $S = (\mathscr{T} \cdot \log_{2}(\mathscr{T}) + \mathscr{N} \cdot \log_{2}(\mathscr{N}))$. 
\end{enumerate}

\begin{algorithm}[ht]
\caption{Uncertainty Quantification using multi-model ensemble: An evolutionary strategies based approach.}
\label{alg:init-uq-algorithm}
\textbf{Input}: Dataset(s) : $\mathcal{D}_T = $ \{$I_i,\; l_i$\}, $\mathcal{D}_V = $ \{$I_i,\; l_i$\}\\
\textbf{Parameters}: \{$ \alpha,\; p,\; N_{epochs}, \;R_{max}^{\mathcal{D}_T }$\}\\
\textbf{Output}: Set of trained Neural Networks : \{$\mathcal{N}_i$\}\\
\begin{algorithmic}[1] 
\STATE Let $t=0$.
\STATE $\mathcal{M}_S[w_t]$ : Randomly Initialized Reference Model.
\WHILE{$t \leq N_{epochs}$}
\STATE \textbf{Perturb and Evaluate}
\FOR{$i$ \textbf{ranging} [1, $p$]}
\STATE \textbf{Using Equation~\ref{weight_perturb_eq}}
\STATE Form pair of perturbed models : $\mathcal{M}[w_{p_i}^{\pm}]$
\STATE \textbf{Obtain performance metrics}
\STATE Evaluate $\mathcal{M}[w_{p_i}^{\pm}]$ on $\mathcal{D}_T$ and $\mathcal{D}_V$
\STATE Store evaluations : $R_i^\pm$ and $_VR_i^\pm$
\ENDFOR
\STATE \textbf{Store models for deep ensemble}
\IF {($R_i^\pm$ == $R_{max}^{\mathcal{D}_T }$) and ($_VR_i^\pm$ \textbf{is unique})}
\STATE Save Model : $\mathcal{M}[w_{p_i}]$
\ENDIF
\STATE \textbf{Update Reference Model}
\STATE \textbf{Using Equation~\ref{Eq_1_3}}
\STATE Compute normalized metrics : $^N R_i^\pm$
\STATE \textbf{Using Equation~\ref{spsa}}
\STATE Compute gradient update for reference model
\STATE $w_{t+1}$ = $w_{t} - \alpha\times\nabla_{w_t} \mathcal{R}$ 
\ENDWHILE
\STATE \textbf{return} Saved Models
\end{algorithmic}
\end{algorithm}

Algorithm~\ref{alg:init-uq-algorithm} more formally defines the process by which models of unique neural schemes are saved during the optimization process. Note that a baseline criteria for model selection with respect to the training set performance is used as the minimum requirement. Additionally the \textbf{is unique} criteria in \textit{line 14} of Algorithm~\ref{alg:init-uq-algorithm} is set such that temporally unique models would be saved. This enables us to scale the ensemble constituents beyond the sparse set of unique validation performance values. For unseen data, the ensemble of the saved models obtained using Algorithm~\ref{alg:init-uq-algorithm} is used to estimate the predictive distribution. The same is used to estimate epistemic uncertainty of the predicted label.

\section{Experimental Setup}

We evaluate the algorithm in a limited data setting where the class of images has inherently low data availability to emphasize the method's necessity. Namely, we employ the method for the detection of choroidal metastases from radiological image slices obtained through Magnetic Resonance Imaging (MRI) of the brain. We used T2-weighted images because they provide adequate contrast between the hyperintense aqueous humor and the lower intense metastatic lesions. 

We aimed to assess the performance of our method relative to the widely used Dropout-based uncertainty quantification (UQ) technique employing autograd ~\cite{gal2016dropout,lakshminarayanan2017simple}. Accordingly, we computed uncertainties using both methods. The results of the Deep NeuroEvolution (DNE)-based UQ are presented herein, while the details of the Dropout-based UQ performance can be found in Appendix A.

\subsection{Data Acquisition and Availability}
This was an IRB-approved study. The data used in this study was acquired while maintaining the highest privacy standards possible and accounting for all ethical considerations. 

We retrospectively obtained MRI brain images from patients with radiology report texts describing choroidal metastases for our tumor-containing cases. We also obtained normal MRI brains for our normal cases. For the tumor-containing cases, we chose T2-weighted images. We selected the axial slice that best exemplified the lesions. Additionally, we manually cropped every image to focus on the right and left globes specifically.  The custom dataset used for this study is available from the corresponding author upon reasonable request.

\subsection{Outline of Experiments}
\begin{figure*}[t]
\centering
\includegraphics[width=0.8\textwidth]{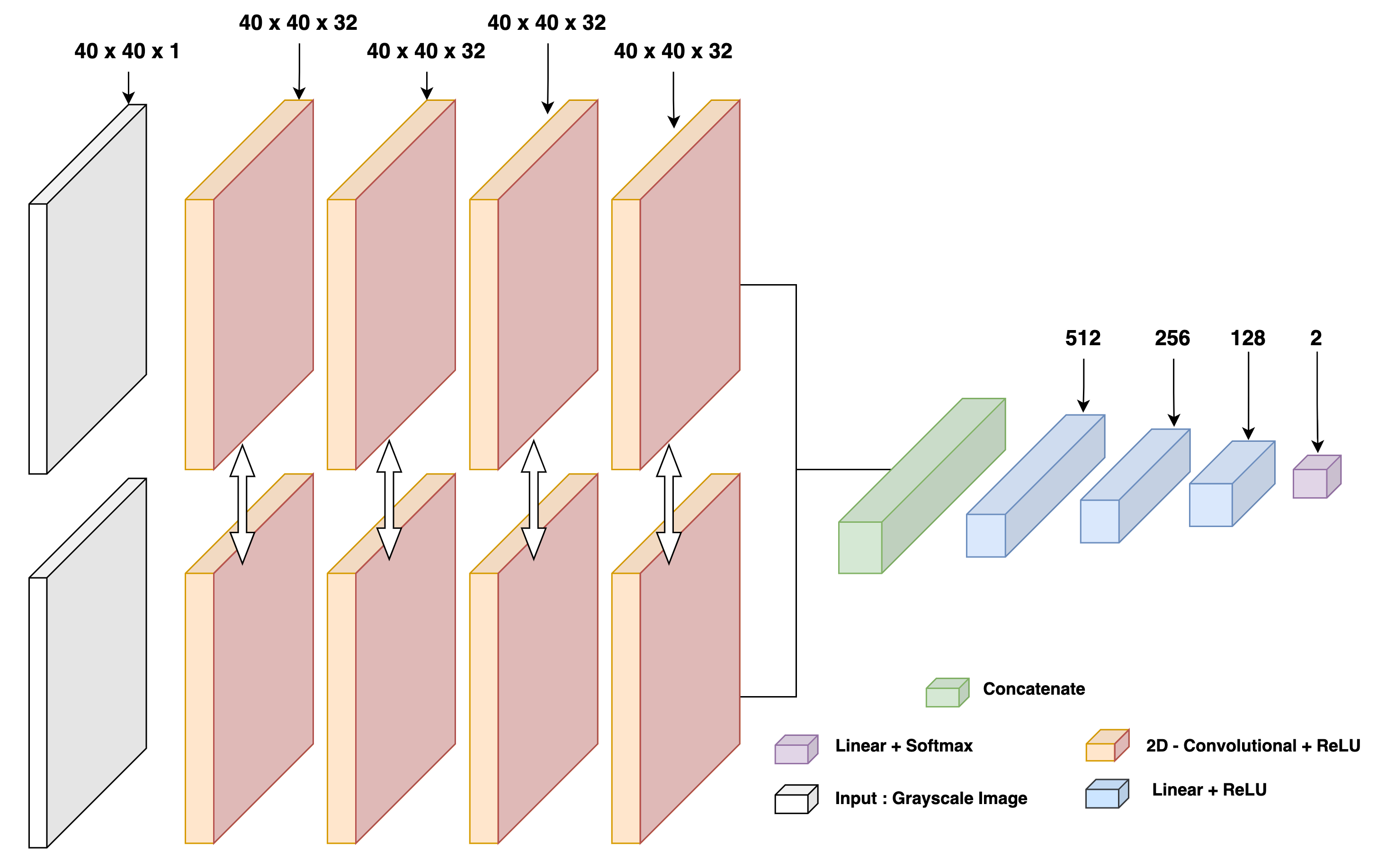} 
\caption{Multi-input convolutional neural network architecture used for the binary classification (tumor vs normal) of choroidal metastases from Brain MRI slices. The network takes as input two grayscale images, each of size (40, 40, 1) : (Height, Width, Channel), and produces a Softmax distribution over the respective classes of interest. Note that the bidirectional arrows represent shared weights between blocks. Respective convolutional channels (for CNN layers) and Neuron counts (Linear layers) are marked against each block. Respective hyperparameters used in training the network as per Algorithm~\ref{alg:init-uq-algorithm} : \{$ \alpha = 0.12,\; p = 40,\; N_{epochs} = 10^5, \;R_{max}^{\mathcal{D}_T } = 30\; or \; 36 $  \}}
\label{arch_2}
\end{figure*}

We evaluated DNE's ability to learn from training data, generalize to a separate testing set, and quantify uncertainty under two distinct neural network settings. First, we tried a single-branched architecture (Figure ~\ref{arch_1}) where only a candidate image slice serves as input. The motivation for doing so was to help the network prioritize inter-class pattern recognition. Second, we employed a two-input (branched) architecture (Figure ~\ref{arch_2}) to facilitate symmetric pattern recognition between left and right globes. Figures~\ref{arch_1} and \ref{arch_2} illustrate these convolutional neural network (CNN) architectures. All model parameters were fixed after basic hyperparameter tuning via grid-search~\cite{bergstra2012random}. A detailed set of hyperparameters used are included in the respective figure captions.

\subsection{Data distribution}
We base our experiments on the class-balanced dataset (1:1). We performed two-fold cross-validation: 
\begin{enumerate}
    \item The first fold featured 18 training set image pairs (images of the bilateral orbits) and 15 testing image pairs.
    \item In the second fold, the training and testing data were flipped, for a 15:18 ratio.
\end{enumerate}
These ratios were chosen to challenge our approach more than is traditionally done in deep learning. Doing so emphasizes DNE's ability to learn and generalize from very small training datasets. 

\section{Results}

Primary evaluations are done so as to evaluate the performance of the proposed setting in terms of relative generalizability; the-best performing model acquired a testing set accuracy of 93.33 $\pm$ 3.33 upon model cross-validation.

\begin{table}[ht]
\centering
\resizebox{\columnwidth}{!}{%
\begin{tabular}{|c|c|c|}
\hline\\
\textbf{UQ class}               & Tumor & Normal \\\\  \hline
Correct with low uncertainty     & 9 (27.3\%)                       & 13 (39.4\%)                             \\ \hline
High uncertainty & 17 (51.2\%)                         & 19 (57.8\%)                           \\ \hline
Incorrect with low uncertainty      & 7 (21.2\%)                          & 1 (3.0\%)                           \\ \hline

\end{tabular}%
}
\caption{Distribution of UQ classes. For purposes of evaluation high uncertainty is formally defined here as Shannon entropy greater than 0.2.}
\label{tab:1}
\end{table}

\begin{figure*}[ht]
\centering
\includegraphics[width=0.8\textwidth, height=9cm]{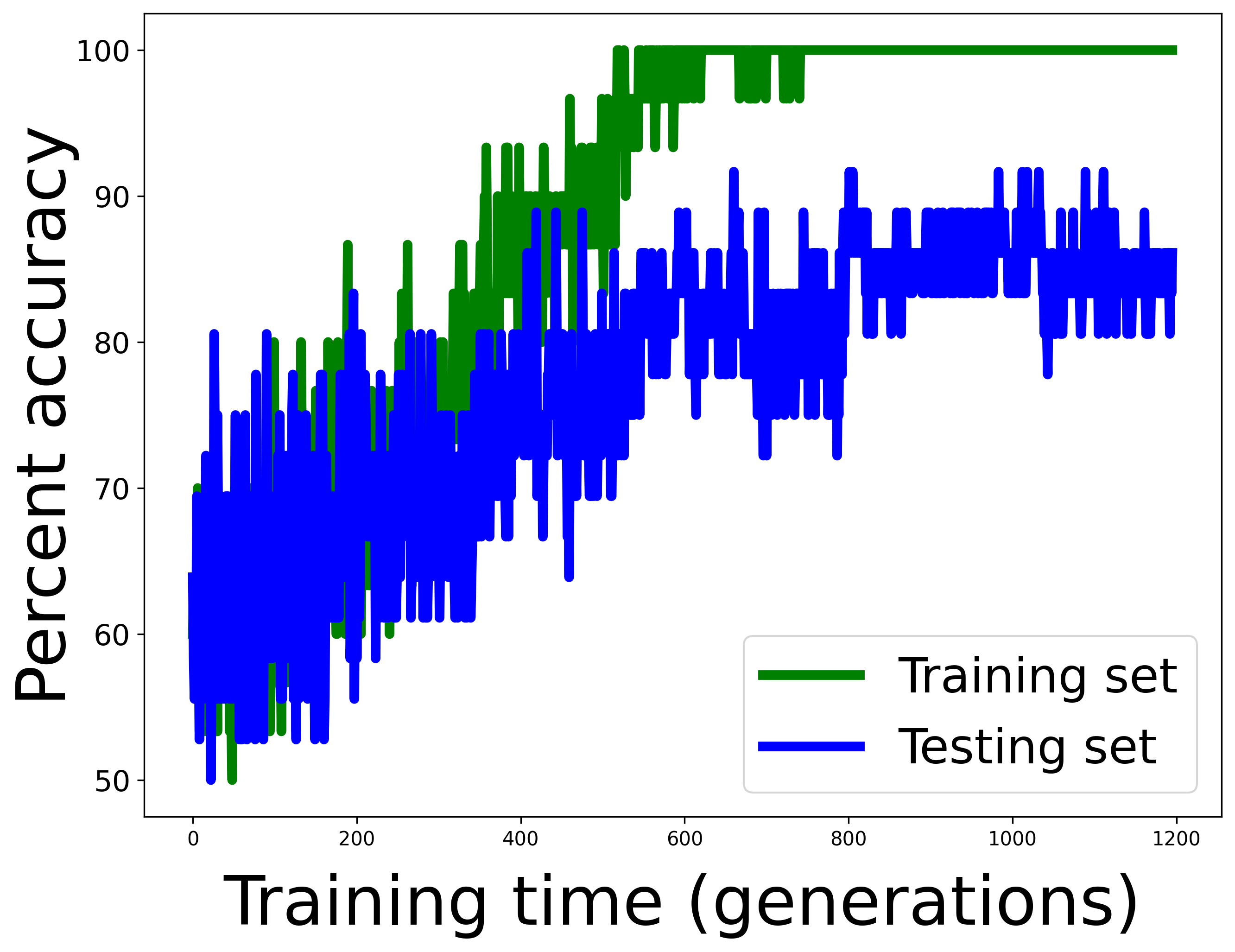} 
\caption{Accuracy versus training time in generations during training of the CNN from Figure \ref{arch_2}. This is an average formed from 5 repeated runs of one of the 2-fold cross-validation training runs, with a training set size of 15 and testing set size of 18.}
\label{acc}
\end{figure*}

\noindent
Next, we acquired predictive distributions of the deep ensemble of models as produced by Algorithm \ref{alg:init-uq-algorithm}. Considering relative entropy as a measure, a subdivision was formed between the predictive distributions so as to form three distinct classes of uncertainties estimated. For the best-performing model out of the two architectures the same is outline in Table~\ref{tab:1}. Figure~\ref{acc} shows the averaged convergence trends shown under evaluations. Further analysis and results on supportive studies are added in Appendix sections A-B. Training and testing code for all models and settings were developed using the Pytorch~\cite{paszke2019pytorch} framework in python. All experiments were conducted on a virtual interface~\cite{bisong2019google} using an 8 core 9th gen intel i7 processor and a Tesla P4 GPU.

\section{Discussion}

\begin{figure*}[t]
\centering
\includegraphics[width=0.8\textwidth]{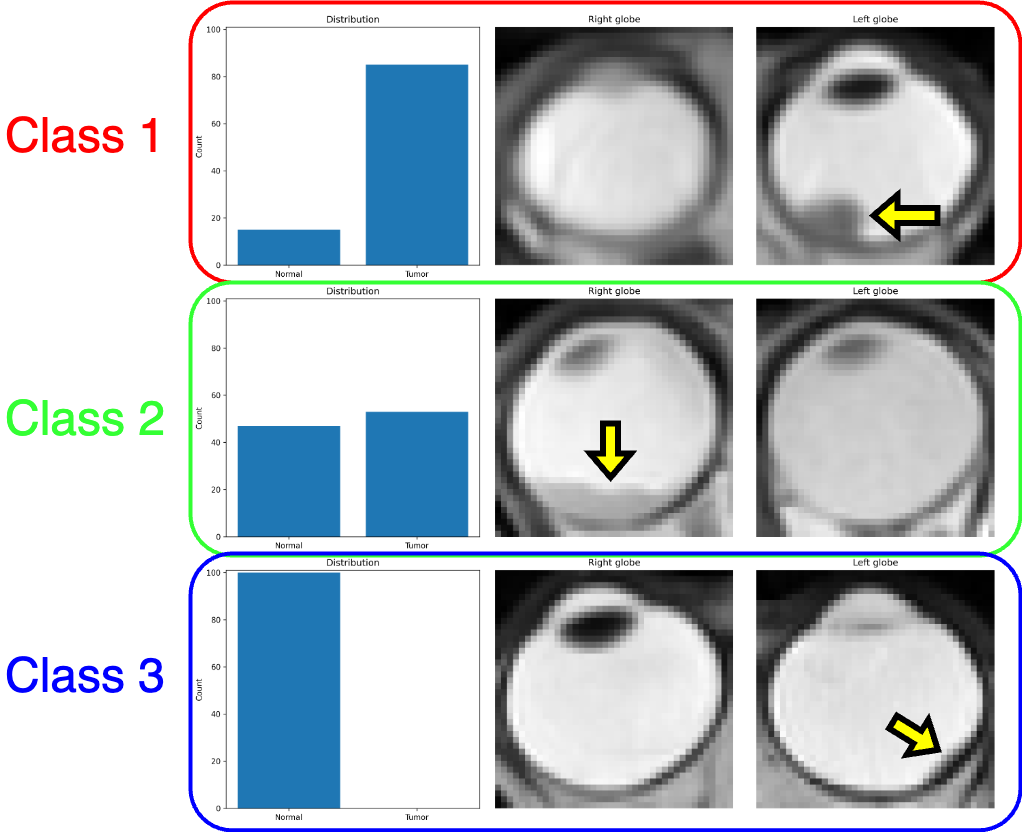} 
\caption{Examples of different uncertainties in tumor-containing cases. The top row displays 
an example from class 1, in which there is a correct prediction with low uncertainty (defined as being less than 20 \%). Class 2, in the middle row, depicts a high-uncertainty scenario. We can see that the tumor is volume-averaged with the adjacent high T2 signal aqueous humor. As such the lesion is less contrastive within the image. Class 3, with an incorrect prediction of normal with low uncertainty, shows a small plaque-like lesion in the left globe that could easily be overlooked by a radiologist.}
\label{tumor_egs_classes}
\end{figure*}

\begin{figure*}[t]
\centering
\includegraphics[width=0.8\textwidth]{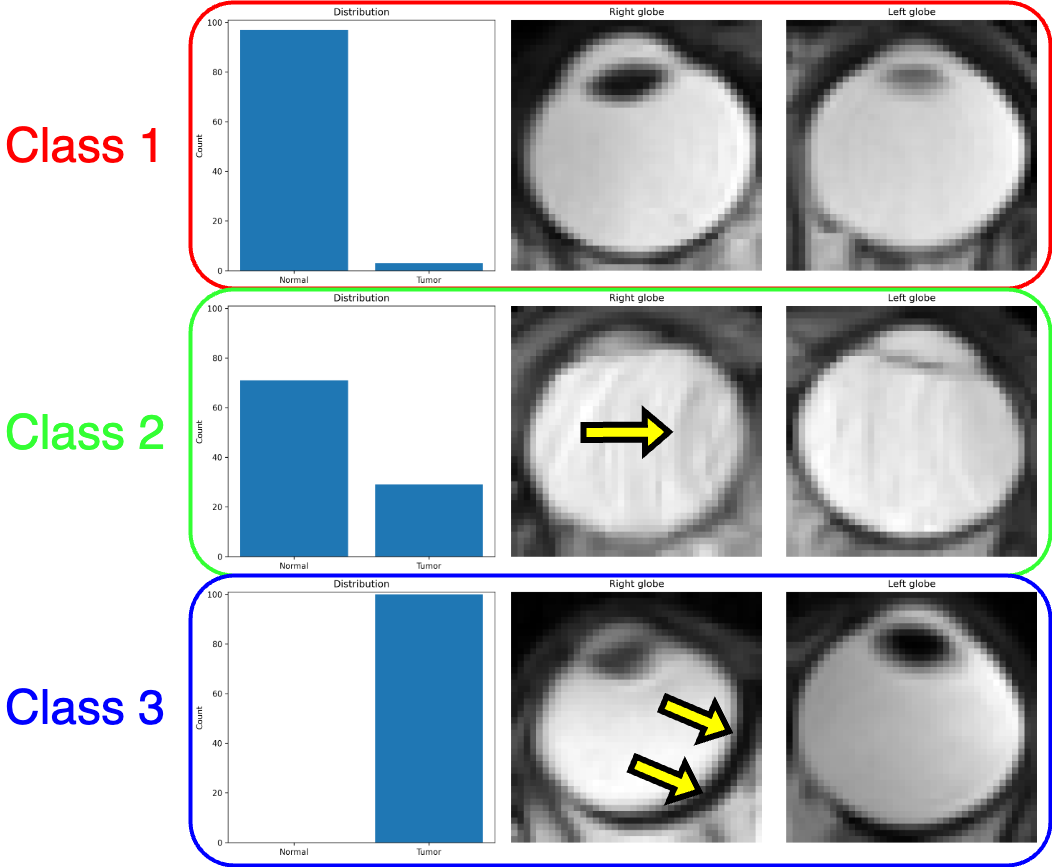} 
\caption{Examples of different uncertainties in normal, tumor-free cases. The top row displays an example from class 1, in which there is a correct prediction with low uncertainty (defined as being less than 20 \%). Class 2, in the middle row, depicts a high-uncertainty scenario. We can see that that the motion and pulsation artifact creates a vague structure that could fool the algorithm into believing it may be a lesion. The class 3 example, which incorrectly predicts a tumor with low unlikelihood, displays apparent thickening at the posterior globe. Although this is due to volume averaging and slice selection, it can be seen to resemble a mass to the AI algorithm.}
\label{normal_egs_classes}
\end{figure*}

Reliable artificial intelligence is of prime importance to medical image analysis because the predictions are high stakes, and trustworthiness is paramount. As such, our models must be able to quantify predictive uncertainty. Quantifying uncertainty has been of interest to many recent studies associated with radiological image analysis~\cite{qendro2021early, abdar2021review, selvan2020uncertainty}. 

Our work expands prior research via the introduction of small data AI techniques and the examination of their relative strengths with respect to quantifying epistemic uncertainty for out-of-distribution data~\cite{yang2021uncertainty, lambert2022trustworthy}. Although the performance factors of ES-based methods~\cite{stember2022deep, ahmadian2021novel, zhou2018sample} on small datasets are impressive and well investigated, this work builds upon the same to provide a means for a seemingly valid uncertainty estimation. Note that a distinguishing factor of our method that sets it apart from prior literature is the guarantee of solution validity for each component of the deep ensemble formed. In other words, on comparison with Monte-Carlo dropout or other forms of deep ensemble formations in an effort to quantify predictive uncertainty, the explicit constraint (as imposed in Algorithm~\ref{alg:init-uq-algorithm}) on the permissible training set accuracy/performance required for each component of the ensemble allows us to form a more valid form of uncertainty modeling. 

From the results outlined in Table~\ref{tab:1}, it is clear that the out-of-distribution data in the testing set largely constitutes the sub-class of relatively high uncertainty and could be identified as possible points of miss-classification. However, it's worth noting that the relative insights that such data points could provide in augmenting the training set are valuable. That's when considering a continual learning model, that requires periodic augmentation of it's training set so as to improve generalizability, the proposed setting could be used as straightforward mechanism for relative selection of data-points from unseen data. 
It is worth mentioning that the vast majority of UQ studies are focused on estimating epistemic uncertainty~\cite{abdar2022hercules, cifci2023deep}, which reflects model uncertainty related to not having seen the needed size and range of data during training to generalize enough to perform on unseen data. Other forms of uncertainty, such as Aleatoric uncertainty, which relates to inherent randomness, are less explored. Note that a practical extension of the method outlined could be to form a set of model communities emerging from different sets of labeling sources (such as institutions or experts) in an effort to quantify other prominent forms of uncertainties. However, in this proof of concept, such depths were not explored.

\section{Limitations}

As previously mentioned in the discussions, there are cases where uncertainty quantification methods could miss out completely while estimating epistemic uncertainty for unseen data. Figures~\ref{tumor_egs_classes} and ~\ref{normal_egs_classes} illustrate the core limitations that come about due to the limitations in the visible feature space of the images or the inherent noise features which could potentially show up in the practical implementation of the method outlined in this study. Another major limitation of the method is the space complexity required to store the distinct models to form the deep ensembles. In future work, we hope to address the above-mentioned shortcomings with better-optimized model configurations~\cite{alibrahim2021hyperparameter} or by relatively simplifying the space complexity using model quantization or quantization aware training~\cite{stewart2021optimising, ferianc2021effects}.

\bibliography{aaai24}

\appendix

\section{Appendix A: Comparison to Monte Carlo Dropout}

\begin{figure*}[ht]
\centering
\includegraphics[width=0.6\textwidth, height=5cm]{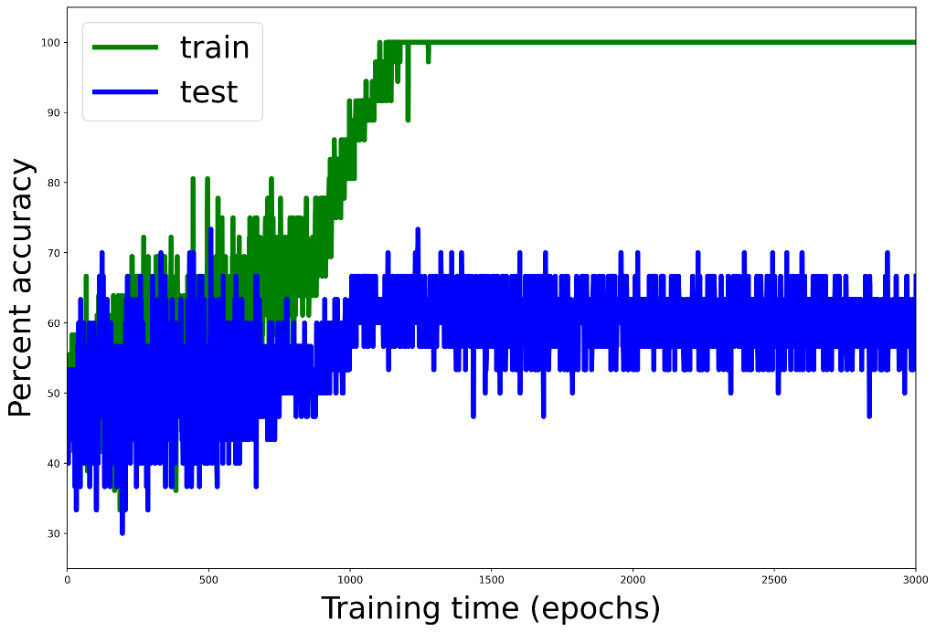} 
\caption{Learning curve during training of the CNN with gradient-based optimization and dropout (Monte Carlo approach). }
\label{acc_monte}
\end{figure*}

Within this supplementary segment, our exploration centers on contrasting Uncertainty Quantification (UQ) employing Deep Neuroevolution (DNE) with the methodology rooted in the Monte Carlo approach. The latter technique encompasses the utilization of a gradient-based optimizer for the training of CNNs), accompanied by the incorporation of dropout in both the training phase and the subsequent uncertainty estimation process. To start, we used essentially the same CNN architecture as for DNE, but we added a dropout layer for both branches after the first fully connected layers. We used a dropout probability of 0.5. We used stochastic gradient descent for optimization, with a learning rate of 0.001. Our objective function was the cross entropy loss. We trained the CNN for 5,000 epochs, after which convergence to 100\% training set accuracy was achieved. As for DNE, we did two-fold cross-validation, with roughly even training:testing splits. Figure~\ref{acc_monte} shows the learning curve for one of the cross-validations. In particular, we note that the variance in testing set accuracy decreases from early during training to later when the system has converged. We can contrast this for DNE’s analogous curve (Figure~\ref{acc}), in which the variations in testing set accuracy also vary less than initially but with higher variance than for the Monte Carlo approach. This reduced variance is due to the overfitting that occurs during training with gradient-based methods and is reflected in Table~\ref{tab:2}. The table shows that, compared to DNE, Monte Carlo has a relative decrease in Class 2 (High uncertainty) even though there is a relatively large prediction error.

\begin{table}[ht]
\centering
{%
\begin{tabular}{|c|c|c|}
\hline
\textbf{UQ class}               & Tumor & Normal \\ \hline
Correct with low uncertainty     & 16 (48.4\%)                      & 9 (27.3\%)                            \\ \hline
High uncertainty & 6 (18.2\%)                         & 6 (18.2\%)                           \\ \hline
Incorrect with low uncertainty      & 11 (33.3\%)                         & 18 (54.5\%)                       \\ \hline

\end{tabular}%
}
\caption{Distribution of UQ classes. For purposes of evaluation, high uncertainty is formally defined here as Shannon entropy greater than 0.2.}
\label{tab:2}
\end{table}

\section{Appendix B: Single-branch CNN architecture}

\begin{figure*}[ht]
\centering
\includegraphics[width=0.8\textwidth]{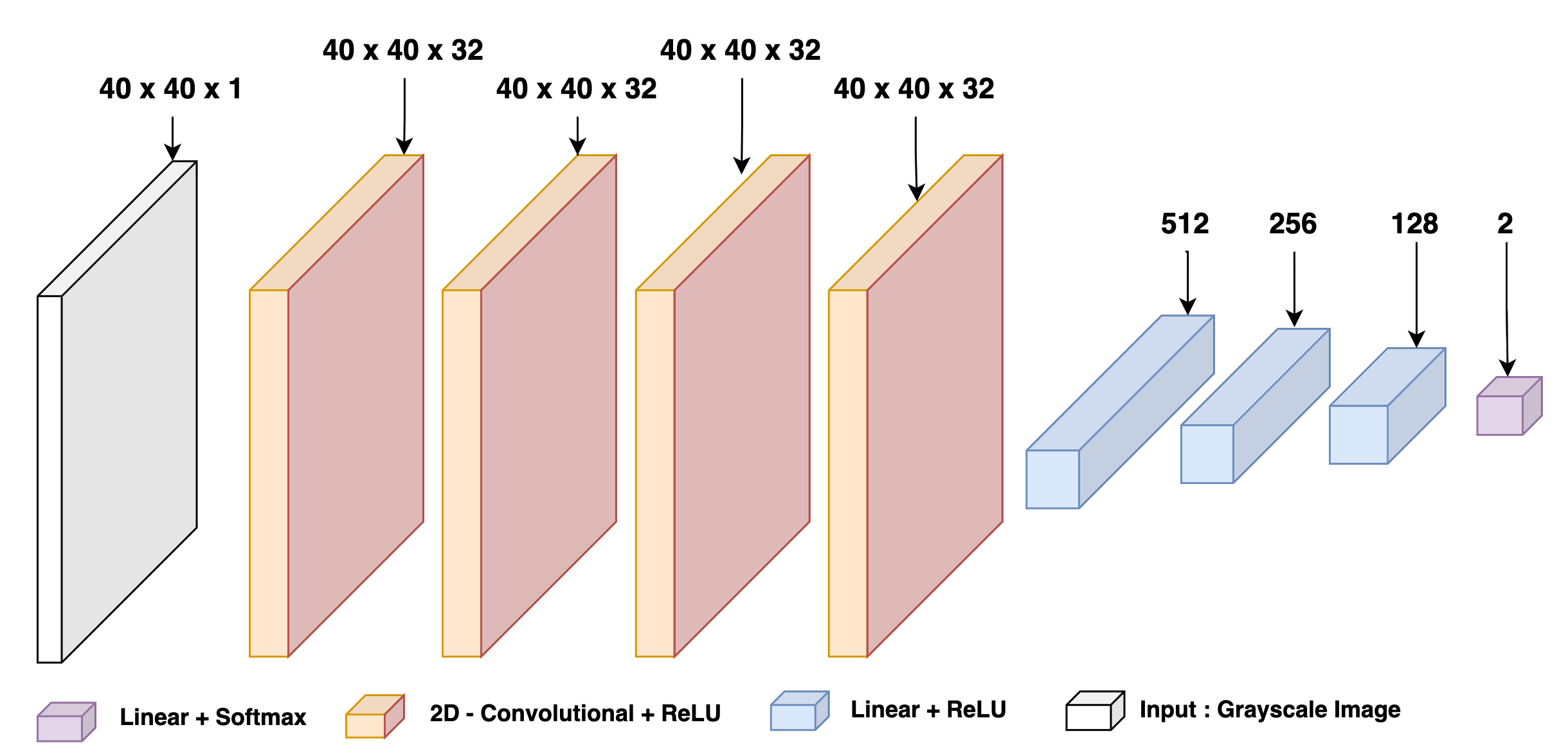} 
\caption{Single input neural network architecture used for the binary classification (tumor vs normal) of choroidal metastases from Brain MRI slices. The network takes as input a grayscale image of size (40, 40, 1) : (Height, Width, Channel) and produces a Softmax distribution over the respective classes of interest. Respective convolutional channels (for CNN layers) and Neuron counts (Linear layers) are marked against each block. Respective hyperparamters used in training the network as per Algorithm~\ref{alg:init-uq-algorithm} : \{$ \alpha = 0.12,\; p = 40,\; N_{epochs} = 10^5, \;R_{max}^{\mathcal{D}_T } = 30\; or \; 36 $  \}}
\label{arch_1}
\end{figure*}

Part of our initial experimentation with various hyperparameter settings involved trying different network architectures. In particular, we initially studied the problem using a single-branch CNN, that took as input single orbital images (Figure ~\ref{arch_1}). Ultimately, the branched architecture of Figure ~\ref{arch_2} performed better and we used the latter for UQ. We can rationalize this by the branched network's added emphasis on symmetry between the orbits; in addition to extracting features of choroidal metastatic lesions, the dual branhced scheme finds features that specifically reflect symmetry or breaking of symmetry. We studied cases in which only one orbit had metastatic disease. The choice reflects the fact that the vast majority of patients with choroidal metastatic disease have it in one eye only. Only one patient had bilateral choroidal metastases and this patient was thus excluded from our study. Given the unilaterality of choroidal metastases, symmetry breaking is an important hallmark of the condition. 

\end{document}